\documentclass[conference,a4paper]{IEEEtran}
\IEEEoverridecommandlockouts
 
\usepackage[hidelinks]{hyperref}
\usepackage[cmex10]{amsmath}
\usepackage{amssymb,amsfonts}
\usepackage{dblfloatfix}
 
\usepackage[ruled,vlined]{algorithm2e}
\usepackage{graphicx}
\graphicspath{{figure/}{Figures/PDF/}{Figures/PNG/}}
 
\usepackage{booktabs}
\usepackage{siunitx}
\DeclareSIUnit{\pixel}{pixel}
\DeclareSIUnit{\percentagepoint}{percentage\ point}
\usepackage[numbers,compress]{natbib}
\usepackage{texnames}
\usepackage{bm,bbm}
\usepackage{orcidlink}
 
\begin{document}
 
\title{Comparison of Deep Learning Frameworks For Rice Disease Mapping From UAV Multispectral Imaging}
 
\author{	\IEEEauthorblockN{Yadav\ Raj\ Ghimire\orcidlink{0009-0002-4446-4176}}
	\IEEEauthorblockA{\textit{North Carolina A\&T State University}\\
		Greensboro, North Carolina, USA.\\
		yrghimire@aggies.ncat.edu}

    \and 
       \IEEEauthorblockN{Jagrati Talreja\orcidlink{0009-0009-4652-4196}}
	\IEEEauthorblockA{\textit{North Carolina A\&T State University}\\
		Greensboro, North Carolina, USA.\\
		jtalreja@ncat.edu}
        
    \and
	\IEEEauthorblockN{Tewodros Syum Gebre\orcidlink{0000-0003-4508-2700}}
	\IEEEauthorblockA{\textit{North Carolina A\&T State University}\\
		Greensboro, North Carolina, USA.\\
		tsgebre@ncat.edu}
        \and

    \and
	\IEEEauthorblockN{Timothy Agboada\orcidlink{0009-0007-6505-9088}}
	\IEEEauthorblockA{\textit{North Carolina A\&T State University}\\
		Greensboro, North Carolina, USA.\\
		tagboada@aggies.ncat.edu}

            \and
	\IEEEauthorblockN{Shikha V. Chandel\orcidlink{0009-0000-7592-6009}}
	\IEEEauthorblockA{\textit{North Carolina A\&T State University}\\
		Greensboro, North Carolina, USA.\\
		svchandel@aggies.ncat.edu}
        \and

           \and
	\IEEEauthorblockN{Leila Hashemi Beni\orcidlink{0000-0003-1026-4555}}
	\IEEEauthorblockA{\textit{North Carolina A\&T State University}\\
		Greensboro, North Carolina, USA.\\
		lhashemibeni@ncat.edu}

}

\maketitle
\begin{abstract}
In this study, UAV multispectral imagery is used to segment the severity of bacterial leaf blight (BLB) in rice using convolutional neural networks (CNNs) and transformer-based models. The evaluated architectures include U-Net with a ResNet-101 encoder, U-Net++ with EfficientNet-B3 and EfficientNet-B7, DeepLabV3+, and SegFormer, all trained under a common pipeline with three input configurations (multispectral only, multispectral+NDVI, and multispectral+NDRE). Experiments are conducted using the publicly available BLB dataset with performance reported using mean IoU (mIoU), mean F1 (mF1), mean accuracy (mAcc), precision, and recall. U-Net++ with EfficientNet-B3 achieved the highest performance, with an mIoU of 97.62\%. SegFormer obtained lower segmentation accuracy but comparable inference speed. Overall, the results indicate that lightweight CNN backbones remain more reliable for operational BLB monitoring while integration of vegetation indices provides small and consistent improvements. The study also highlights the value of standardised UAV datasets to compare disease mapping methods and encourages the use of CNN architectures for field implementation.
 
\end{abstract}
 
 
\section{Introduction}
 
Precision agriculture depends on rapid and accurate disease mapping as rice diseases affect food security and farmer income \cite{ismail2025comprehensive}. Deep learning has emerged as a leading tool for early identification of plant disease, demonstrating significant progress over traditional methods \cite{ismail2025comprehensive,yusuf2024systematic}. U-Net models can be improved with attention mechanisms, giving good segmentation results in remote sensing \cite{jamali2024residual}. Encoder–decoder architectures such as U-Net \cite{ronneberger2015u} and U-Net++ \cite{zhou2018unet++} are widely adopted for remote sensing segmentation. Recent studies use these models for crop monitoring with aerial imagery to map rice fields \cite{zhang2025uav} and other fields. CNN designs like EfficientNet and DeepLabV3+, which combine atrous spatial pyramid pooling with an encoder-decoder structure for precise boundary recovery \cite{chen2018encoder}, and architectures like SegFormer \cite{dong2025segformer} continue to deliver dependable results in dense prediction tasks \cite{sharma2025dba}. Deep learning has been successfully applied to rice BLB detection using UAV multispectral imagery \cite{dorbu2024detection}.

Unmanned aerial vehicles (UAVs) fitted with multispectral or hyperspectral sensors allow monitoring early signs of disease more efficiently than field inspections \cite{zhang2025uav}. This feature has been further expanded by coordinated multi-UAV systems, which allow for efficient and methodical coverage of wide agricultural areas with shorter mission times \cite{gray2025multi}. Bands like the red-edge and near-infrared (NIR) detect changes in leaf health not visible to the eye. Vegetation indices such as NDVI and NDRE capture these changes \cite{maimaitijiang2025estimating}, enabling modern deep learning models to perform semantic segmentation directly on patches of a multispectral image. This approach is more consistent for recognizing complex visual patterns of disease than relying on a single calculated index \cite{zhang2025uav}.
 
U-Net introduced an encoder-decoder design with skip links for image segmentation \cite{ronneberger2015u}. U-Net++ subsequently introduced nested skip paths to enhance multi-scale feature integration \cite{zhou2019unet++}. ResNet backbones further enable deeper feature learning through residual connections and have been widely integrated into U-Net, DeepLab, and related architectures. EfficientNet has also implemented composite scaling to balance the size and precision of the model  \cite{tan2019efficientnet}, which is also the encoder. A comparison of encoder–decoder designs including U-Net, SegNet, FCN, and DeepLabV3+ for crop–weed discrimination in UAS imagery found that DeepLabV3+ with a ResNet-18 backbone obtained the best weed extraction performance \cite{hashemi2022deep}. Recently, transformer-based encoders such as the MIT series in SegFormer can model long-range dependencies in remote sensing data and have demonstrated strong results in segmentation \cite{xie2021segformer}. The use of vision transformer models across sensor types in remote sensing has been confirmed by implementing them to instance segmentation of LiDAR point clouds for environmental mapping applications \cite{yang2023instance}. However, few studies compare these encoder–decoder families on the same agricultural UAV dataset despite their proven effectiveness in other crop-disease segmentation settings \cite{dorbu2021uav}.

Some studies evaluated different CNN architectures for rice BLB segmentation and demonstrated that U-Net with ResNet101 encoder trained on NDVI multispectral data can achieve mIoU \SI{97.20}{\percent} and effectively distinguish healthy rice from low and high-severity BLB \cite{logavitool2025field}.  However, these studies tested a limited set of models and did not include transformer-based architectures. This study addresses that gap by providing the first controlled benchmarking of five encoder–decoder architectures including U-Net++, DeepLabV3+, and SegFormer on the BLB multispectral dataset and evaluated across the three spectral input configurations and a unified training protocol. Therefore, results provide evidence-based guidance for architecture selection in operational rice disease monitoring systems. The value of standardised publicly available aerial datasets for such benchmarking has been demonstrated across remote sensing domains, including high-resolution flood mapping \cite{fawakherji2025deepflood} and integrated LiDAR--multispectral environmental monitoring \cite{anokye2026integrating}.

\section{Materials and Methods}
 
\subsection{Study area and UAV data}
 
This experiment uses a publicly available UAV multispectral BLB dataset \cite{logavitool2025field} which collected over the rice fields in Thailand. The imagery was captured using a Phantom~4 Multispectral UAV drone which was flown at approximately \SI{20}{\meter} altitude. The camera captured blue, green, red, red-edge, and NIR bands with \SI{1.1}{\centi\meter\per\pixel} spatial resolution and with \SI{70}{\percent} overlap. Disease severity information was provided within the dataset and converted to pixel-level masks representing healthy rice, low-severity BLB, high-severity BLB, and background classes such as soil, water, and weeds. Orthomosaics and label maps were available with the dataset.

\subsection{Band combinations and dataset variants}
 
Following three input configurations are considered:
\begin{itemize}
    \item \textbf{D1 (Multispectral)}: blue, green, red, red-edge, NIR.
    \item \textbf{D2 (Multispectral + NDVI)}: D1 plus NDVI computed from red and NIR.
    \item \textbf{D3 (Multispectral + NDRE)}: D1 plus NDRE computed from red-edge and NIR.
\end{itemize}
For each configuration, the selected bands are stacked into six-channel images and normalized to $[0,1]$ using min--max scaling.
 
\subsection{Patch extraction and data split}
 
Orthomosaics and label maps are tiled into overlapping $256\times256$~pixel patches with $128\times128$~pixel overlap. Zero-padding is applied where necessary to ensure full coverage. The dataset is divided into training, validation, and test subsets in a 70:15:15 split, consistent with producing roughly 1,299 training, 280 validation, and 280 test images. Data augmentation (random flips, rotations, and brightness changes) is applied to training patches only.
 
\subsection{Segmentation Models}
 
This study evaluates four encoder--decoder architectures using identical datasets and training settings:
\begin{itemize}
    \item \textbf{U-Net--ResNet101}: baseline U-Net model with a ResNet101 encoder.
    \item \textbf{U-Net++--EfficientNet-B3}: a U-Net++ variant using an EfficientNet-B3 encoder.
    \item \textbf{U-Net++--EfficientNet-B7}: a larger EfficientNet encoder variant.
    \item \textbf{DeepLabV3+--ResNet101}: encoder-decoder architecture using atrous separable convolution  with a ResNet-101 backbone.
    \item \textbf{SegFormer (MiT-B2)}: a transformer-based encoder with a lightweight segmentation head.
\end{itemize}
 
All models are trained on a workstation equipped with dual NVIDIA RTX~A6000 GPUs (\SI{48}{GB} each). Differences in available compute compared with earlier experiments allow larger batch sizes and faster convergence, which can contribute to small performance variations. Models are implemented in PyTorch using the \texttt{segmentation\_models\_pytorch} library where applicable, with support for six-channel multispectral input.

\subsection{Training protocol}
 
All experiments use the same training setup:
\begin{itemize}
    \item Optimizer: Adam with initial learning rate $1\times10^{-4}$.
    \item Batch size: 16 patches.
    \item Maximum epochs: 500.
    \item Loss: a weighted sum of categorical cross-entropy and soft Dice loss to balance class frequencies and emphasize boundary quality.
    \item Early stopping: based on validation mIoU with a patience of 30 epochs.
\end{itemize}
 
For each model and band configuration the checkpoint with the highest validation mIoU is kept and evaluated on the held-out test set.

\subsubsection*{Training Loss Convergence}
\begin{figure}[t]
    \centering
    \includegraphics[width=0.45\textwidth]{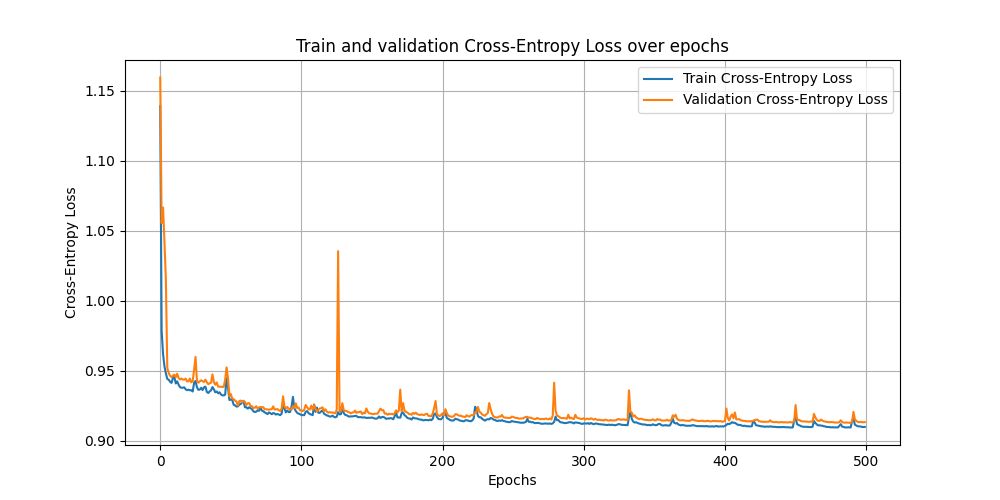}
    \caption{Training and validation loss curves over 500 epochs using the combined categorical cross-entropy and soft Dice loss.}
    \label{fig:loss_curve}
\end{figure}

To monitor the learning behaviour, the combined loss was recorded on both the training and validation sets. As shown in Fig.~\ref{fig:loss_curve}, the loss decreases steadily over the course of training, indicating stable convergence without major overfitting. This confirms that the weighted cross-entropy and soft Dice formulation is appropriate for the multispectral disease segmentation task.
 
\subsection{Evaluation metrics}
 
Let $p_{ij}$ denote the number of pixels belonging to ground-truth class $i$ that are predicted as class $j$. For class $k$, define:
\begin{align}
\mathrm{IoU}_k &= 
\frac{p_{kk}}{\sum_j p_{kj} + \sum_i p_{ik} - p_{kk}}, \\
\mathrm{Acc}_k &= 
\frac{p_{kk}}{\sum_j p_{kj}}, \\
\mathrm{F1}_k &= 
\frac{2\,\mathrm{Prec}_k\,\mathrm{Rec}_k}{\mathrm{Prec}_k + \mathrm{Rec}_k},
\end{align}
where $\mathrm{Prec}_k$ and $\mathrm{Rec}_k$ are precision and recall for class $k$. This study reports mean IoU (mIoU), mean pixel accuracy (mAcc), mean F1-score (mF1), mean precision (mPrec), and mean recall (mRec) averaged over the four semantic classes. Inference time is measured as the average time to process a batch of 16 patches on the GPU.
 
 
\section{Experimental Results}
\label{sec:results}
 
\subsection{Overall comparison on D2}
 
Table~\ref{tab:comparison_d2} lists the statistical output of all evaluated models on the D2 configuration (multispectral + NDVI). All models achieve very high accuracy with mIoU values above \SI{95}{\percent} for CNN-based architectures. Among them U-Net++ with EfficientNet-B3 achieves the best overall comparison, slightly surpassing the U-Net--ResNet101 baseline across all metrics.
 
\begin{table}[t]
\centering
\caption{Model comparison on the D2 dataset (multispectral + NDVI).}
\label{tab:comparison_d2}
\scriptsize
\setlength{\tabcolsep}{1.5pt}
\resizebox{\columnwidth}{!}{%
\begin{tabular}{l l S[table-format=2.2] S[table-format=2.2] S[table-format=2.2] S[table-format=2.2] S[table-format=2.2] S[table-format=2.2]}
\toprule
\multicolumn{2}{c}{Model} &
\multicolumn{1}{c}{mIoU} &
\multicolumn{1}{c}{mAcc} &
\multicolumn{1}{c}{mF1} &
\multicolumn{1}{c}{mPrec} &
\multicolumn{1}{c}{mRec} &
\multicolumn{1}{c}{Inf.} \\
Arch. & Backbone &
\multicolumn{1}{c}{(\%)} &
\multicolumn{1}{c}{(\%)} &
\multicolumn{1}{c}{(\%)} &
\multicolumn{1}{c}{(\%)} &
\multicolumn{1}{c}{(\%)} &
\multicolumn{1}{c}{(it/s)} \\
\midrule
U-Net      & ResNet-101      & 97.07 & 99.39 & 98.50 & 97.83 & 99.18 & 18.83 \\
DeepLabV3+ & ResNet-101      & 95.75 & 99.10 & 97.79 & 96.84 & 98.77 & 17.85 \\
U-Net++    & EfficientNet-B3 & \textbf{97.62} & \textbf{99.51} & \textbf{98.79} & \textbf{98.23} & \textbf{99.35} & 18.83 \\
U-Net++    & EfficientNet-B7 & 97.84 & 99.56 & 98.90 & 98.41 & 99.40 & 13.50\\
SegFormer  & MiT-B2         & 90.36 & 97.85 & 94.86 & 91.48 & 98.53 & 18.83 \\
\bottomrule
\end{tabular}%
}
 
\end{table}

U-Net++--EfficientNet-B3 achieves an mIoU of \SI{97.62}{\percent}, mAcc of \SI{99.51}{\percent}, and mF1 of \SI{98.79}{\percent}. The classical model U-Net--ResNet101 baseline is only slightly less (mIoU \SI{97.07}{\percent}) which confirms that a deep residual encoder remains a reliable choice for this experiment. DeepLabV3+--ResNet101 is approximately \SI{1.3}{\percentagepoint} lower in mIoU than U-Net++--EfficientNet-B3 but still remains high overall accuracy. SegFormer--MiT-B2 lags in mIoU (\SI{90.36}{\percent}) and precision but very high recall, indicating a tendency to identify the diseased areas. Additionally, inference throughput in the RTX~A6000 is between about 17--19~images/s for most of the models but with U-Net++--EfficientNet-B7 expected to be slower (around 13--14~images/s) due to its heavier encoder.
 
\subsection{Qualitative segmentation results}
 
\begin{figure*}[t]
    \centering
    \includegraphics[width=\textwidth]{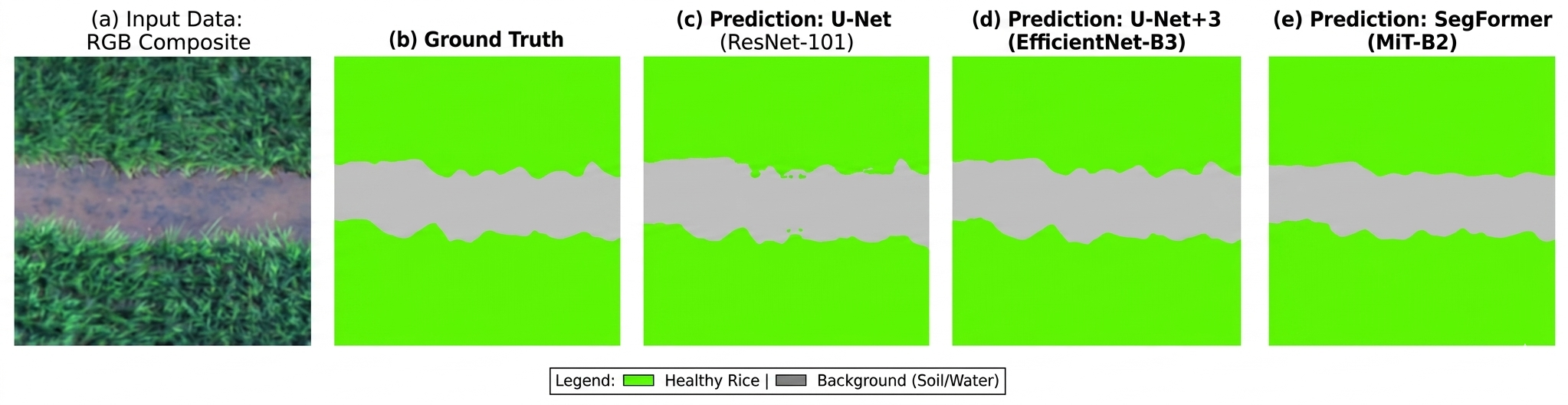}
    \caption{Spatial segmentation comparison on a D2 test patch: (a) RGB composite from multispectral imagery, (b) ground truth annotation, (c) U-Net--ResNet101 prediction, (d) U-Net++--EfficientNet-B3 prediction, and (e) SegFormer--MiT-B2 prediction. Green indicates healthy rice vegetation, while gray represents background classes including soil and water.}
    \label{fig:spatial_results}
\end{figure*}

Figure~\ref{fig:spatial_results} presents spatial segmentation maps comparing the top-performing models on representative test patches from the D2 dataset. The visualization shows: (a) the original multispectral RGB composite, (b) ground truth annotation, (c) U-Net--ResNet101 prediction, (d) U-Net++--EfficientNet-B3 prediction, and (e) SegFormer--MiT-B2 prediction. U-Net++--EfficientNet-B3 demonstrates superior boundary delineation between healthy rice and background regions, with predictions closely matching the ground truth mask. The model accurately captures the spatial pattern of the water channel separating rice fields. All three models produce visually similar segmentation results in this example, with clear distinction between healthy vegetation and background classes. These visual comparisons confirm that CNN-based architectures provide reliable spatial outputs for operational disease mapping, with U-Net++--EfficientNet-B3 offering the best balance of accuracy and computational efficiency as demonstrated in Table~\ref{tab:comparison_d2}.
 
\subsection{Performance across band combinations D1--D3}
 
The CNN-based models and SegFormer were trained on D1 (multispectral image only), D2 (multispectral + NDVI) and D3 (multispectral + NDRE). Table~\ref{tab:combos} reports the metrics for each dataset combination. For U-Net and U-Net++--EfficientNet-B3 performance is very consistent across D1--D3 with small gains on D3 for U-Net++.
 
\begin{table}[t]
\centering
\caption{Effect of band combinations (D1, D2, D3) on segmentation performance.}
\label{tab:combos}
\scriptsize
\setlength{\tabcolsep}{2pt}
\begin{tabular}{l l l S[table-format=2.2] S[table-format=2.2] S[table-format=2.2]}
\toprule
Dataset & Arch. & Backbone & {mIoU} & {mAcc} & {mF1} \\
       &       &          & {(\%)} & {(\%)} & {(\%)} \\
\midrule
D1 & U-Net    & ResNet-101      & 96.88 & 99.35 & 98.39 \\
D2 & U-Net    & ResNet-101      & 97.07 & 99.39 & 98.50 \\
D3 & U-Net    & ResNet-101      & 97.26 & 99.43 & 98.59 \\
\midrule
D1 & U-Net++  & EfficientNet-B3 & 97.61 & 99.51 & 98.77 \\
D2 & U-Net++  & EfficientNet-B3 & 97.62 & 99.51 & 98.79 \\
D3 & U-Net++  & EfficientNet-B3 & \textbf{97.73} & \textbf{99.53} & \textbf{98.84} \\
\midrule
D1 & U-Net++  & EfficientNet-B7 & 97.80 & 99.55 & 98.88 \\
D2 & U-Net++  & EfficientNet-B7 & 97.84 & 99.56 & 98.90 \\
D3 & U-Net++  & EfficientNet-B7 & 97.76 & 99.54 & 98.86 \\
\midrule
D1 & SegFormer & MiT-B2         & 89.12 & 97.43 & 94.05 \\
D2 & SegFormer & MiT-B2         & 90.36 & 97.85 & 94.86 \\
D3 & SegFormer & MiT-B2         & 89.12 & 97.43 & 94.05 \\
\bottomrule
\end{tabular}
\end{table}
 
Several patterns emerge:
 
\begin{itemize}
\item For U-Net--ResNet101, mIoU improves slightly from D1 (\SI{96.88}{\percent}) to D3 (\SI{97.26}{\percent}) which suggesting that NDRE shows a tiny increment in accuracy.
\item For U-Net++--EfficientNet-B3, all three combinations perform almost the same but with a slightly high accuracy in D3 (mIoU \SI{97.73}{\percent}). This shows that the model can extract reliable features from either NDVI or NDRE augmented stacks.
\item SegFormer-MiT-B2 is consistently weaker across all dataset with mIoU between \SI{89}{\percent} and \SI{90}{\percent} even though mAcc and mRec remain high.
\end{itemize}

\subsection{Inference throughput}
 
Table~\ref{tab:throughput} summarizes inference speed on the RTX~A6000, measured as images per second for batches of 16 patches. All CNN-based models process more than 17~images/s, while SegFormer is comparable. U-Net++--EfficientNet-B7 is the slowest model due to its larger encoder, but still fast enough for near-real-time mapping on a single GPU.
 
\begin{table}[t]
\centering
\caption{Inference throughput on D2 (batch size 16, RTX~A6000).}
\label{tab:throughput}
\begin{tabular}{l l S[table-format=2.2]}
\toprule
Architecture & Backbone & {Inf. (it/s)} \\
\midrule
U-Net        & ResNet-101      & 18.83 \\
DeepLabV3+   & ResNet-101      & 17.85 \\
U-Net++      & EfficientNet-B3 & 18.83 \\
U-Net++      & EfficientNet-B7 & 13.89 \\
SegFormer    & MiT-B2          & 18.83 \\
\bottomrule
\end{tabular}
\end{table}

\section{Discussion}
 
The studies verify that decoder and encoder selections significantly affect BLB segmentation performance with a fixed preprocessing pipeline and training protocol. U-Net++ with EfficientNet-B3 consistently achieves the highest mIoU and mF1 across all datasets. Similarly, DeepLabV3+--ResNet101 achieves comparable accuracy but is about approximately \SI{1.3}{\percentagepoint} below the U-Net++--EfficientNet-B3 in mIoU in the D2 dataset model. Similarly for SegFormer, which uses the backbone MiT-B2 clearly shows the underperformance compared to the CNN-based models in mIoU (around \SIrange{89}{90}{\percent} across D1--D3). Its recall is very high (\textgreater\SI{98}{\percent}) but precision is lower, leading to systematic overestimation of diseased areas. This suggests that transformer-based architectures may be ill-suited for relatively small multispectral datasets without targeted regularization or domain-specific pretraining and which may require careful regularization, data augmentation, or pretraining tailored to remote sensing. The indices like NDVI or NDRE improves segmentation accuracy as compared to single multispectral bands. These performance trends are consistent with prior findings showing improved metrics upon addition of a vegetation index channel. In this experiment, NDRE yields marginally higher accuracy  than NDVI across most configurations. Therefore, selecting an appropriate vegetation index is a meaningful factor in maximizing segmentation performance.
Table~\ref{tab:throughput} shows that all CNN-based models can process a large number of patches per second on a single RTX~A6000. U-Net++--EfficientNet-B3 achieves the best balance of accuracy and inference speed whereas the B7 variant shows minor improvements at the expense of efficiency. SegFormer provides similar throughput but substantially weaker accuracy. Therefore, CNN backbones are currently the most reliable option for operational BLB mapping.

\section{Conclusion}
 
The experiments highlight that U-Net++--EfficientNet-B3 emerges as the most reliable architecture for this dataset with stable performance with different band configurations. Stacking multispectral bands with a vegetation index channel (NDVI or NDRE) is sufficient to reach mIoU above \SI{97}{\percent}, enabling reliable canopy-level disease mapping. The accuracy and speed of U-Net++--EfficientNet-B3 is a promising model for deployment in the real field. These findings emphasize the importance of the publicly available UAV rice disease datasets such as the BLB collection. By using the same dataset and pipeline while varying only the model architecture, this study provides clearer evidence of which deep learning models are best suited for crop disease detection. This work is limited to a single field season and multispectral data. In the future studies, this work will extend across regions and years and integrate with the thermal or LiDAR sensor for earlier stress detection.

\section*{Acknowledgment}
THIS WORK IS SUPPORTED BY NASA AWARD 80NSSC23M0051 AND NSF AWARD 2401942.
 
\small
\bibliographystyle{IEEEtranN}
\bibliography{references}
 
\end{document}